\documentclass[runningheads]{llncs}
\usepackage{makeidx}
\usepackage{graphicx}
\usepackage{amsmath,amssymb} 
\usepackage{color}

\usepackage{booktabs}
\usepackage{subcaption}
\usepackage{bm}

\newcommand{\etal}{\emph{et al}.~}
\newcommand{\ie}{\emph{ie}.~}

\begin{document}
	\title{Temporal Interpolation as an Unsupervised Pretraining Task for Optical Flow Estimation}

    \titlerunning{Temporal Interpolation for Optical Flow Pretraining}

\author{Jonas Wulff\inst{1,2} \quad \quad
Michael J. Black\inst{1}}
%
	\authorrunning{Wulff, J. and Black, M.J.}
%

\institute{Max-Planck Institute for Intelligent Systems, T\"{u}bingen, Germany \and
  MIT CSAIL, Cambridge, MA, USA \\
  \email{wulff@mit.edu, black@tuebingen.mpg.de}}
	\maketitle

	\begin{abstract}
	The difficulty of annotating training data is a major obstacle to using CNNs for low-level tasks in video. Synthetic data often does not generalize to real videos, while unsupervised methods require heuristic losses. Proxy tasks can overcome these issues, and start by training a network for a task for which annotation is easier or which can be trained unsupervised. The trained network is then fine-tuned for the original task using small amounts of ground truth data.
	Here, we investigate frame interpolation as a proxy task for optical flow. Using real movies, we train a CNN unsupervised for temporal interpolation. Such a network implicitly estimates motion, but cannot handle untextured regions. By fine-tuning on small amounts of ground truth flow, the network can learn to fill in homogeneous regions and compute full optical flow fields. Using this unsupervised pre-training, our network outperforms similar architectures that were trained supervised using synthetic optical flow.

\end{abstract}

	\section{Introduction}
In recent years, the successes of deep convolutional neural networks (CNNs) have led to a large number of breakthroughs in most fields of computer vision.
The two factors that made this possible are a widespread adoption of massively parallel processors in the form of GPUs
and large amounts of available annotated training data.
To learn good and sufficiently general representations of visual features, CNNs require tens of thousands to several hundred million~\cite{Sun:2017:UnreasonableEffectiveness} examples of the visual content they are supposed to process, annotated with labels teaching the CNNs the desired output for a given visual stimulus.

\newcommand{\teaserwidth}{0.9\textwidth}
\begin{figure}
	\centering
	\includegraphics[width=\teaserwidth]{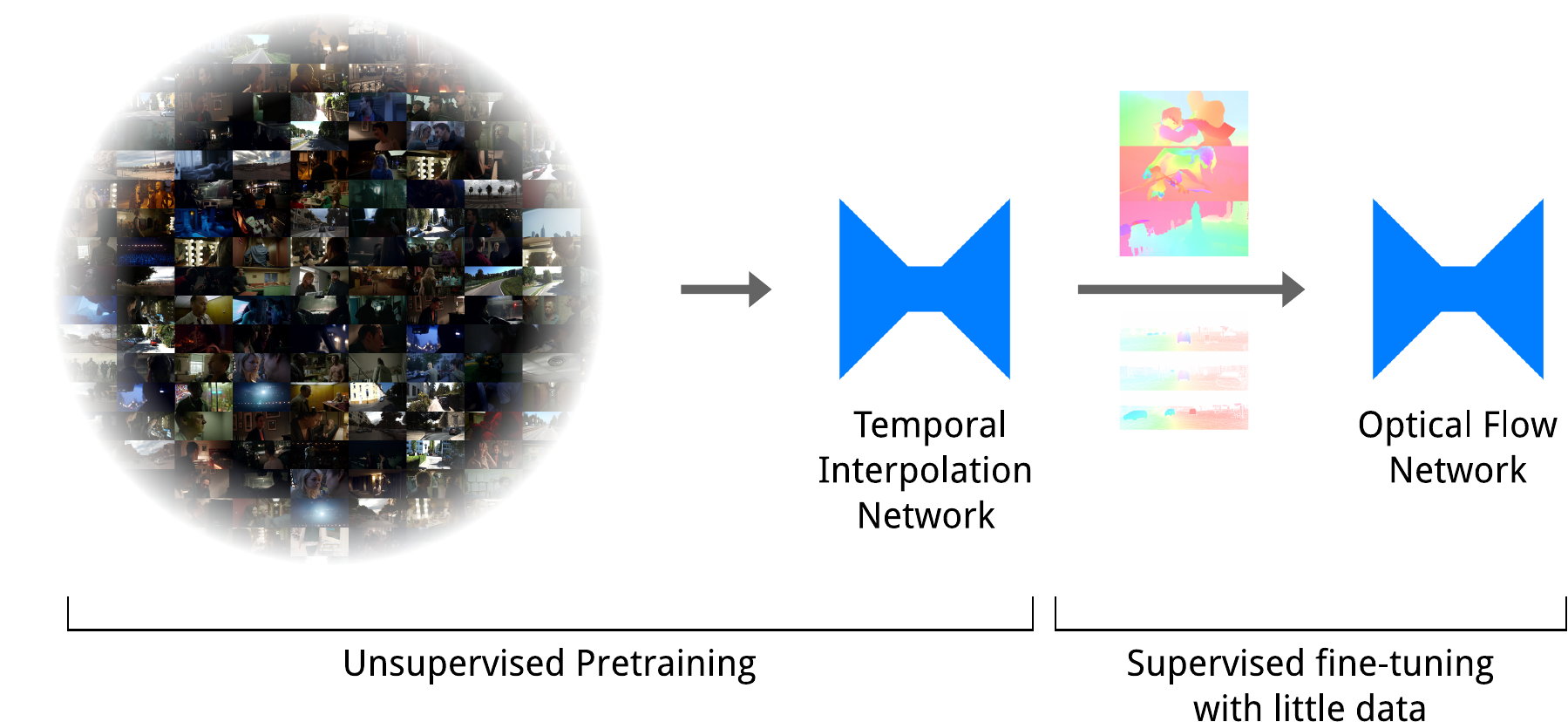}
	\caption{\small Overview of our method. Using large amounts of unlabelled data we train a temporal interpolation network without explicit supervision. We then fine-tune the network for the task of optical flow using small amounts of ground truth data, outperforming similar architectures that were trained supervised.}
	\label{fig:teaser}
\end{figure}

For some tasks such as image classification or object detection it is possible to generate massive amounts of data by paying human workers to manually annotate images. For example, writing a description of an image into a textbox or dragging a rectangle around an object are tasks that are easy to understand and relatively quick to perform.
For other tasks, especially those related to video, obtaining ground truth is not as easy.
For example, manual annotation of dense optical flow, motion segmentation, or tracking of objects requires not just the annotation of a huge number of instances (in the first two cases one would ideally want one annotation per pixel), but an annotator would also have to step back and forth in time to ensure temporal consistency~\cite{Liu:2008:HumanAssistedMotionAnnotation}. 
Since this makes the task both tedious and error-prone, manually annotated data is rarely used for low-level video analysis.

An alternative is to use synthetic data, for example from animated movies~\cite{Butler:ECCV:Sintel}, videogames~\cite{Richter:2017:PlayingForBenchmarks}, or generated procedurally \cite{Mayer:2016:FlyingThings}.
The problem here is that synthetic data always lives in a world different from our own.
Even if the low-level image statistics are a good match to those of the real world, it is an open question whether realistic effects such as physics or human behavior can be learned from synthetic data.

A different approach to address the issue of small data is to use \textit{proxy tasks}.
The idea is to train a network using data for which labels are easy to acquire, or for which unsupervised or self-supervised learning is possible.
Training on such data forces the network to learn a latent representation, and if the proxy task is appropriately chosen, this representation transfers to the actual task at hand.
The trained network, or parts thereof, can then be fine-tuned to the final task using limited amounts of annotated data, making use of the structure learned from the proxy task.
An example proxy task for visual reasoning about images is colorization, since solving the colorization problem requires the network to learn about the semantics of the world~\cite{Larsson:2017:ColorizationPretraining}.
Another example is context prediction~\cite{Doersch:2015:ContextPrediction}, in which the proxy task is to predict the spatial arrangement of sub-patches of objects, which helps in the final task of object detection.

What, then, would be a good proxy task for video analysis?
One core problem in the analysis of video is to compute temporal correspondences between frames.
Once correspondences are established, it is possible to reason about the temporal evolution of the scene, track objects, classify actions, and reconstruct the geometry of the world.
Recent work by Long \etal~\cite{Long:2016:LearningImageMatching} proposed a way to learn about correspondences without supervision.
They first train a network to interpolate between two frames.
For each point in the interpolated frame, they then backpropagate the derivatives through the network, computing which pixels in the input images most strongly influence this point.
This, in turn, establishes correspondences between the two maxima of these derivatives in the input images.
Their work, however, had two main shortcomings.
First, using a complete backpropagation pass to compute each correspondence is computationally expensive.
Second, especially in unstructured or blurry regions of the frame, the derivatives are not necessarily well located, since a good (in the photometric sense) output pixel can be sampled from a number of wrongly corresponding sites in the input images; frame interpolation does not need to get the flow right to produce a result with low photometric error.
This corresponds to the classical aperture problem in optical flow, in which the flow is not locally constrained, but context is needed to resolve ambiguities.
Consequently, so far, the interpolation task has not served as an effective proxy for learning flow.

In this work, we address these shortcomings and show that, treated properly, the interpolation task can, indeed, support the learning of optical flow.
To this end, we treat training for optical flow as a two-stage problem.
In the first stage, we train a network to estimate the center frame from four adjacent, equally spaced frames.
This forces the network to learn to establish correspondences in visually distinct areas.
Unlike previous work, which used only limited datasets of a few tens of thousands frames such as KITTI-RAW~\cite{Geiger2013:KITTI}, we use a little under one million samples from a diverse set of datasets incorporating both driving scenarios and several movies and TV series.
This trains the network to better cope with effects like large displacements and motion and focus blur.
Thanks to this varied and large body of training data, our network outperforms specialized frame interpolation methods despite not being tailored to this task.

In a second stage, we fine-tune the network using a small amount of ground-truth optical flow data from the training sets of KITTI~\cite{Geiger2013:KITTI} and Sintel~\cite{Butler:ECCV:Sintel}.
This has three advantages. First, after fine-tuning, the network outputs optical flow directly, which makes it much more efficient than~\cite{Long:2016:LearningImageMatching}.
Second, this fine-tuning forces the network to group untextured regions and to consider the context when estimating the motion; as mentioned above, this can usually not be learned from photometric errors alone.
Third, compared to fully unsupervised optical flow algorithms~\cite{Ahmadi:2016:UnsupervisedCNNForMotion,Meister:2017:UnFlow}, during training our method does not employ prior assumptions such as spatial smoothness, but is purely data-driven.

Our resulting network is fast and yields optical flow results with superior accuracy to the comparable networks of FlowNetS~\cite{Dosovitskiy:2015:FlowNet} and SpyNet~\cite{Ranjan:2016:Spynet} which were trained using large amounts of labeled, synthetic optical flow data~\cite{Ranjan:2016:Spynet}.
This demonstrates that (a) when computing optical flow, it is important to use real data for training, and (b) that temporal interpolation is a suitable proxy task to learn from to make learning from such data feasible.
	\section{Previous work}

\textbf{CNNs for Optical Flow.}
In the past years, end-to-end training has had considerable success in many tasks of computer vision, including optical flow.
The first paper demonstrating end-to-end optical flow was FlowNet~\cite{Dosovitskiy:2015:FlowNet}, which used an architecture similar to ours, but trained on large amounts of synthetic ground truth optical flow.
In follow-up work~\cite{Ilg:2017:Flownet2}, the authors propose a cascade of hourglass networks, each warping the images closer towards each other. Furthermore, they significantly extend their training dataset (which is still synthetic).
This leads to high performance at the cost of a complicated training procedure.

Taking a different direction, SpyNet~\cite{Ranjan:2016:Spynet} combines deep learning with a spatial pyramid.
Similar to classical optical flow methods, each pyramid level computes the flow residual for the flow on the next coarser scale, and successively warps the input frame using the new, refined flow.
This allows the authors to use a very simple network architecture, which in turns leads to high computational efficiency.
The training, however, is still done using the same annotated training set as~\cite{Dosovitskiy:2015:FlowNet}.
The recently proposed PWC-Net~\cite{Sun:2017:PWC} uses ideas of both, and computes the flow in a multiscale fashion using a cost volume on each scale, followed by a trained flow refinement step.

A different approach is to not train a network for full end-to-end optical flow estimation, but to use trained networks inside a larger pipeline.
Most commonly, these approaches use a network to estimate the similarity between two patches~\cite{Guney:2016:DeepDiscreteFlow,Xu:2017:DCFlow}, effectively replacing the data cost in a variational flow method by a CNN.
The resulting data costs can be combined with spatial inference, for example belief-propagation~\cite{Guney:2016:DeepDiscreteFlow}, or a cost volume optimization~\cite{Xu:2017:DCFlow}.
A network trained to denoise data can also be used as the proximal operator in a variational framework~\cite{Meinhardt:2017:LearningProximalOperators}.
In the classical optical flow formulation, this would correspond to a network that learns to regularize.

All these approaches, however, require large amounts of annotated data.
For real sequences, such training data is either hard to obtain for general scenes~\cite{Janai:2017:SlowFlow}, or limited to specific domains such as driving scenarios~\cite{Geiger2013:KITTI}.
Synthetic benchmarks~\cite{Butler:ECCV:Sintel,Mayer:2016:FlyingThings,Richter:2017:PlayingForBenchmarks}, on the other hand, often lack realism, and it is unclear how well their training data generalizes to the real world.

Hence, several works have investigated unsupervised training for optical flow estimation.
A common approach is to let the network estimate a flow field, warp the input images towards each other using this flow field, and measure the similarity of the images under a photometric loss.
Since warping can be formulated as a differentiable function~\cite{Jaderberg:2015:STN}, the photometric loss can be back-propagated, forcing the network to learn better optical flow.
In~\cite{Yu:2016:BackToBasics}, the authors combine the photometric loss with a robust spatial loss on the estimated flow, similar to robust regularization in variational optical flow methods~\cite{Sun:2014:IJCV}.
However, while their training is unsupervised, they do not demonstrate cross-dataset generalization, but train for Flying Chairs~\cite{Dosovitskiy:2015:FlowNet} and KITTI~\cite{Geiger2013:KITTI} using matching training sets and separately tuned hyper-parameters.
In~\cite{Ren:2017:UnsupervisedOpticalFlowEstimation}, the authors use the same approach, but show that a network that is pre-trained using the same dataset as in~\cite{Dosovitskiy:2015:FlowNet} can be fine-tuned to different output scenarios.
Similarly, USCNN~\cite{Ahmadi:2016:UnsupervisedCNNForMotion} uses only a single training set.
Here, the authors do not use end-to-end training, but instead use a photometric loss to train a network to estimate the residual flow on different pyramid layers, similar to~\cite{Ranjan:2016:Spynet}.
The recently proposed UnFlow~\cite{Meister:2017:UnFlow} uses a loss based on the CENSUS transform and computes the flow in both forward and backward direction. This allows the authors to integrate occlusion reasoning into the loss; using an architecture based on FlowNet2, they achieve state-of-the-art results on driving scenarios.
All these methods require manually chosen heuristics as part of the loss, such as spatial smoothness or forward-backward consistency-based occlusion reasoning. Therefore, they do not fully exploit the fact that CNNs allow us to overcome such heuristics and to purely ``let the data speak''.
In contrast, our method does not use any explicitly defined heuristics, but uses an unsupervised interpolation task and a small number of ground truth flow fields to learn about motion.

Several approaches use geometrical reasoning to self-supervise the training process.
In TransFlow~\cite{Alletto:2017:TransFlow}, the authors train two networks, a first, shallow one estimating a global homography between two input frames, and a second, deep network estimating the residual flow after warping with this homography.
Since they use the homography to model the ego-motion, they focus on driving scenarios, and do not test on more generic optical flow sequences.
In~\cite{Godard:2017:UnsupervisedDepthEstimation}, a network is trained to estimate depth from a single image, and the photometric error according to the warping in a known stereo setup induced by the depth is penalized. Similarly~\cite{Zhou:2017:UnsupervisedDepthAndEgoMotion} trains a network to estimate depth from a single image by learning to warp, but use videos instead of stereo.
SfM-Net~\cite{Fragkiadaki:2017:SfMNet} learns to reduce a photometric loss by estimating the 3D structure of the world, the motion of the observer, and the segmentation into moving and static regions, which is enough to explain most of the motion of the world. However, as most other methods, it is only tested on the restricted automotive scenario.

Simliar to self-supervision using geometric losses, Generative Adversarial Networks~\cite{Goodfellow:2014:GAN} have been used to learn the structure of optical flow fields. In~\cite{Lai:2017:SemiSupervisedFlow}, the GAN is trained to distinguish between the warping errors caused by ground truth and wrongly estimated optical flow. Only the discriminator uses annotated data; once trained, it provides a loss for unsupervised training of the flow itself.

\textbf{Frame interpolation.}
Instead of warping one input frame to another, it is also possible to train networks to interpolate and extrapolate images by showing them unlabeled videos at training time.
A hybrid network is used in~\cite{Sedaghat:2017:NextFramePrediction}, where a shared contractive network is combined with two expanding networks to estimate optical flow and to predict the next frame, respectively.
Similar to us, they hypothesize that temporal frame generation and optical flow share some internal representations; however, unlike us they train the optical flow network completely with labeled data, and do not test on the challenging Sintel test set.
Similarly, \cite{Vondrick:2016:AnticipatingVisualRepresentations} trains a network to anticipate the values of intermediate feature maps in the future.
However, they are not interested in the motion itself, but in the future higher-level scene properties such as objects and actions.
Niklaus \etal~\cite{Niklaus:2017:AdaptiveConvolution} propose a video interpolation network, where the motion is encoded in an estimated convolutional kernel; however, the quality of this implicit motion is never tested.
As mentioned above, the work most similar to ours is~\cite{Long:2016:LearningImageMatching}, where a CNN is trained to interpolate between frames and subsequently used to compute correspondences between images.
Unlike our work, however, they require expensive backpropagation steps to establish the correspondences; we show that using a small amount of training data can teach the network to translate between its internal motion representation and optical flow, resulting in significantly improved performance.
	\section{A frame interpolation network}
\newcommand{\architecturewidth}{0.9\textwidth}
\begin{figure*}[t]
	\centering
	\includegraphics*[width=\architecturewidth]{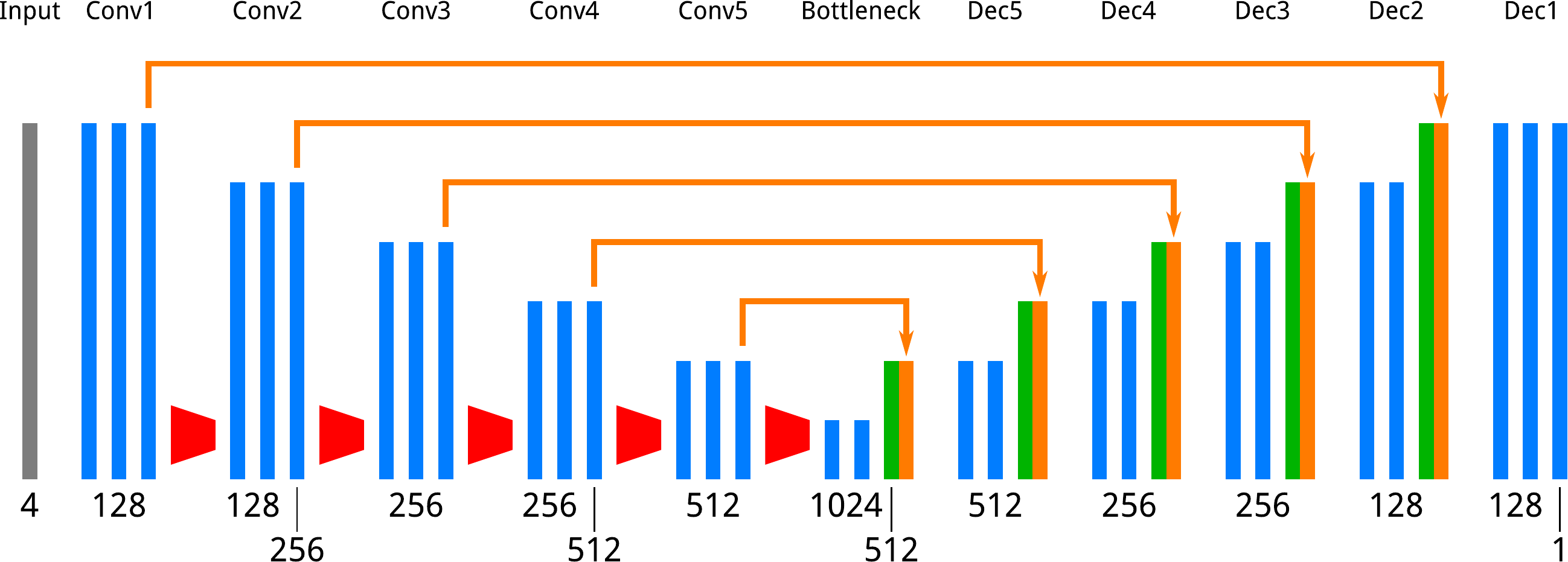}
	\caption{\small Architecture of our network. The outputs of $3 \times 3$ convolutional layers are shown in blue, the output of $2 \times 2$ max-pooling operations in red, $2\times$ transposed convolutions in green, and side-channel concatenation in orange. Not shown are leaky ReLUs after each layer except the last and batch normalization layers. The numbers indicate the number of channels of each feature map.}
	\label{fig:architecture}
\end{figure*}

\newcommand{\spacingwidth}{0.9\textwidth}
\begin{figure}[t]
	\centering
	\includegraphics[width=\spacingwidth]{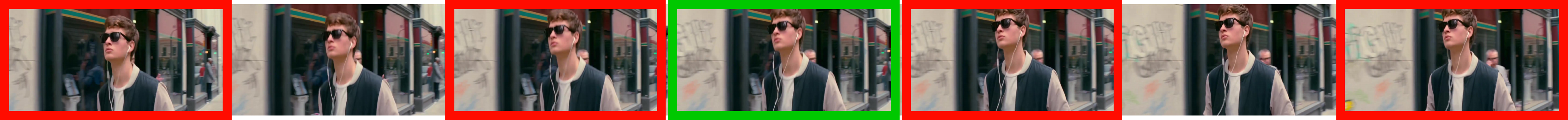}
	\caption{\small Our network predicts the center frame (green) from four neighboring, equally spaced frames (red). To ensure equal spacing of the input frames, the unmarked frames (second from the left and right) are not taken into account.}
	\label{fig:spacing}
\end{figure}

The core hypothesis of our work is that in order to properly perform temporal interpolation, it is necessary to learn about the motion of the world.
Temporal interpolation, however, is a task that does not require explicit supervision; with proper selection of input and output frames, every video sequence can serve as a supervisory signal.
Therefore, we start by training a network for the task of interpolation in an unsupervised manner.
Given four frames (as shown in Fig.~\ref{fig:spacing} in red), the task of our network is to interpolate the center frame, shown in green in Fig.~\ref{fig:spacing}.
Empirically, we found that using four input frames resulted in approximately 13\% lower errors (2.04 px EPE) on the optical flow estimation task than using two frames (2.31 px EPE).
We believe that the reasons for this are that with more than two frames (a) it is easier to reason about higher order temporal effects (\ie non-zero acceleration), and (b) it enables the network to reason about occlusions, which requires at least three frames~\cite{Sun:2012:Layers}.
We hence use four input frames for both the interpolation and the optical flow estimation task.
We use grayscale images as input and output, since we found the final optical flow to have lower errors with grayscale than with color images.

\textbf{Network architecture.}
Similar to~\cite{Long:2016:LearningImageMatching}, we use a standard hourglass architecture with side-channels, as shown in Fig.~\ref{fig:architecture}.
This simple architecture is nevertheless surprisingly effective in many different applications, such as optical flow computation~\cite{Dosovitskiy:2015:FlowNet} and unsupervised depth estimation~\cite{Zhou:2017:UnsupervisedDepthAndEgoMotion}.
Our network consists of five convolutional blocks (\texttt{Conv1} to \texttt{Conv5} in Fig.~\ref{fig:architecture}), each of which contains three $3 \times 3$ convolutional layers followed by batch normalization layers and leaky ReLUs.
Between the blocks, we use max-pooling to reduce the resolution by a factor of two.
Within each block, all layers except the last output the same number of feature maps.
\texttt{Conv5} is followed by a bottleneck block consisting of two convolutional layers and a transposed convolution.
The output is then successively upscaled using a series of decoder blocks (\texttt{Dec5} to \texttt{Dec1}).
Each consists of two convolutional layers and (except for \texttt{Dec1}) a transposed convolution which doubles the resolution, again interleaved with leaky ReLUs and batch normalization layers.

To preserve high frequency information, we use side channels as in~\cite{Dosovitskiy:2015:FlowNet}, shown in orange in Fig.~\ref{fig:architecture}.
The output of the transposed convolutions are concatenated with the appropriate outputs from the convolutional layers.
The last convolutional layer of \texttt{Dec1} directly outputs the monochrome output image and is not followed by a nonlinearity.
Table~\ref{tab:sources} summarizes the number of inputs and outputs of each convolutional block.

\begin{table}	
	\setlength{\tabcolsep}{3pt}
	\centering
	\caption{\small Number of input and output channels per layer block.}	
	\scriptsize
	\begin{tabular}{lccccccccccc}  
		\toprule
		& Conv1 & Conv2 & Conv3 & Conv4 & Conv5 & Bottleneck & Dec5 & Dec4 & Dec3 & Dec2 & Dec1 \\
		\midrule
		Input & 4 & 128 & 128 & 256 & 256 & 512 & 1024 & 1024 & 512 & 512 & 256 \\
		Output & 128 & 128 & 256 & 256 & 512 & 1024 & 1024 & 512 & 512 & 256 & 1 \\
		\bottomrule
	\end{tabular}
	\label{tab:sources}
\end{table}	

\textbf{Training data.}
Unsupervised training would in theory allow us to train a network with infinite amounts of data.
Yet, most previous works only use restricted datasets, such as KITTI-RAW.
Instead, in this work we aim to compile a large, varied dataset, containing both cinematic sequences from several movies of different genres as well as more restricted but very common sequences from driving scenarios.
As movies, we use \textit{Victoria} and \textit{Birdman}. The former was filmed in a true single take, and the later was filmed and edited in such a way that cuts are imperceptible.
In addition, we use long takes from the movies \textit{Atonement}, \textit{Children of Men}, \textit{Baby Driver}, and the TV series \textit{True Detective}.
Shot boundaries would destroy the temporal consistency that we want our network to learn and hence would have to be explicitly detected and removed; using single-take shots eliminates this problem.

For the driving scenarios, we use KITTI-RAW~\cite{Menze2015GCPR} and Malaga~\cite{Blanco2014}, the first of which contains several sequences and the second of which contains one long shot of camera footage recorded from a driving car.
For each sequence, we use around 1 \% of frames as validation, sampled from the beginning, center, and end of the sequence and sampled such that they do not overlap the actual training data.
The only difference is KITTI-RAW, which is already broken up into sequences.
Here, we sample full sequences to go either to the training or validation set, and use 10 \% for the validation set. This ensures that the validation set contains approximately the same amount of frames from driving and movie-like scenarios.
Table~\ref{tab:training_samples} summarizes the datasets used and the amount of frames from each.

In total, we thus have approximately 464K training frames.
However, in movie sequences, the camera and object motions are often small.
Therefore, during both training and validation, we predict each frame twice, once from the adjacent frames as shown in Fig.~\ref{fig:spacing}, and once with doubled spacing between the frames.
Therefore, a target frame at time $t$ provides two training samples, one where it is predicted from the frames at $t-3, t-1, t+1,$ and $t+3$, and one where it is predicted from $t-6, t-2, t+2,$ and $t+6$.
This ensures that we have a sufficient amount of large motions in our frame interpolation training set.
In total, our training and validation sets contain 928,410 and 16,966 samples, respectively.

\textbf{Training details.}
As shown in Fig.~\ref{fig:spacing}, each training sample of our network consists of the four input frames and the center frame which the network should predict.
During training, each quadruple of frames is separately normalized by subtracting the mean of the four input frames and dividing by their standard deviation.
We found this to work better than normalization across the full dataset.
We believe the reason for this is that in correspondence estimation, it is more important to consider the structure \textit{within} a sample than the structure across samples, the later of which is important for classification tasks.
To put it differently, whether a light bar or a dark bar moves to the right does not matter for optical flow and should produce the same output.

As data augmentation, we randomly crop the input images to rectangles with the aspect ratio 2:1, and resize the cropped images to a resolution of $384 \times 192$ pixel, resulting in randomization of both scale and translation.
For all input frames belonging to a training sample, we use the same crop.
Furthermore, we randomly flip the images in horizontal and vertical direction, and randomly switch between forward and backward temporal ordering of the input images.
We use a batch size of 8, train our network using ADAM and use a loss consisting of a structural similarity loss (SSIM)~\cite{Wang:2004:SSIM} and an $L_1$ loss, weighted equally.
The initial learning rate is set to $1e-4$, and halved after 3, 6, 8, and 10 epochs.
We train our network for 12 epochs; after this point, we did not notice any further decrease in our loss on the validation set.
\begin{table}
	\centering
	\caption{\small Training data for interpolation.}	
	\begin{tabular}{cccc}  
		\toprule
		Source & Type & Training frames & Validation frames \\
		\midrule
		Birdman & Full movie & 155,403 & 1,543 \\
		Victoria & Full movie & 188,700 & 1,878 \\
		KITTI-RAW & Driving & 39,032 & 3,960 \\
		Malaga & Driving & 51,285 & 460 \\
		Atonement & Movie clip & 7,062 & 44 \\
		Children of Men & Movie clip & 9,165 & 65 \\
		Baby Driver & Movie clip & 3,888 & 14 \\
		True Detective & Movie clip & 8,388 & 57 \\
		\midrule
		Total & & 464,205 & 8,483 \\
		\bottomrule
	\end{tabular}
	\label{tab:training_samples}
\end{table}

	\section{From interpolation to Optical Flow}
Given a frame interpolation network, it has been shown before~\cite{Long:2016:LearningImageMatching} that motion is learned by the network and can be extracted.
However, this only works for regions with sufficient texture.
In unstructured areas of the image, the photometric error that is used to train the frame interpolation is not informative, and even a wrong motion estimation can result in virtually perfect frame reconstruction.

What is missing for good optical flow estimation, then, is the capability to group the scene and to fill in the motion in unstructured regions, \ie to address the aperture problem.
Furthermore, the mechanism used to extract the motion in~\cite{Long:2016:LearningImageMatching} is slow, since it requires a complete backpropagation pass for each correspondence.
To effectively use frame interpolation for optical flow computation, two steps are thus missing: (a) to add knowledge about grouping and region fill-in to a network that can compute correspondences, and (b) to modify the network to directly yield an optical flow field, making expensive backpropagation steps unnecessary at test time.
Luckily, both objectives can be achieved by fine-tuning the network to directly estimate optical flow, using only a very limited amount of annotated ground truth data.

For this, we replace the last layer of the \texttt{Dec1} block with a vanilla $3 \times 3$ convolutional layer with two output channels instead of one, and train this network using available ground truth training data, consisting of the training sets of KITTI-2012~\cite{Geiger2013:KITTI}, KITTI-2015~\cite{Menze2015GCPR} and the clean and final passes of MPI-Sintel~\cite{Butler:ECCV:Sintel}, for a total of about 2500 frames.
We use 10 \% of the data as validation.

We again use ADAM with an initial learning rate of $10^{-4}$, halve the learning rate if the error on the validation set has not decreased for 20 epochs, and train for a total of 200 epochs using the endpoint error as the loss.
Except for the temporal reversal, we use the same augmentations as described above.

As we will see in the next section, this simple fine-tuning procedure results in a network that computes good optical flow, and even outperforms networks with comparable architecture that were trained using large amounts of synthetically generated optical flow.
	\section{Experiments}
\label{sec:experiments}
In this section, we demonstrate the effectiveness of our method for interpolation and optical flow estimation, and provide further experiments showing the importance of pre-training and the effects of reduced ground truth data for fine-tuning.

\subsection{Temporal interpolation}

\newcommand{\interpolationresult}{0.3\textwidth}
\begin{figure}[t]
	\centering
	\centerline{
		\includegraphics[width=\interpolationresult]{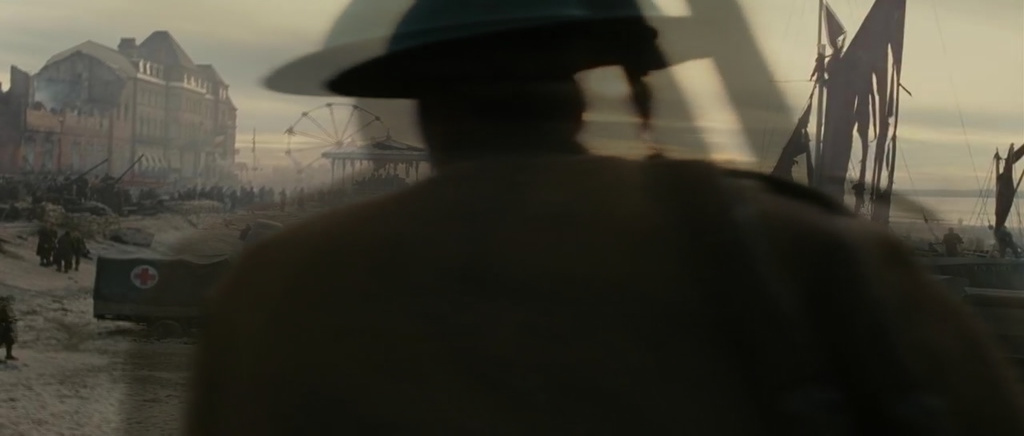}
		\includegraphics[width=\interpolationresult]{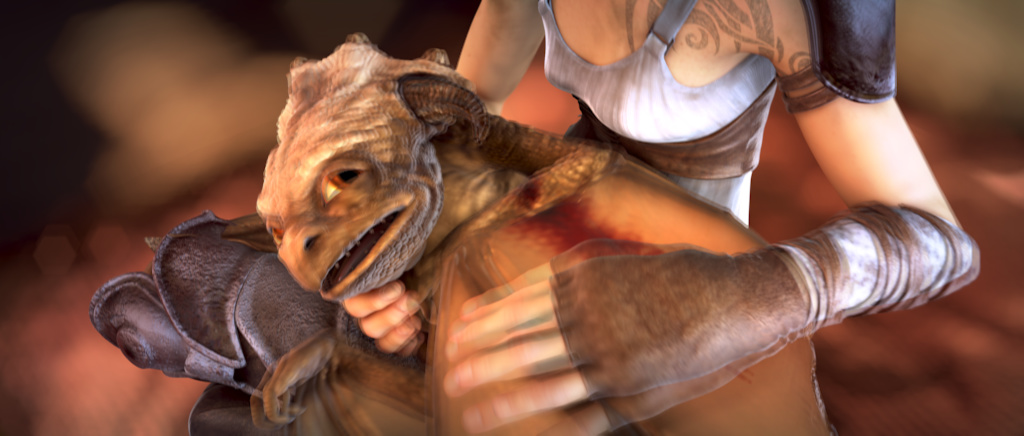}
		\includegraphics[width=\interpolationresult]{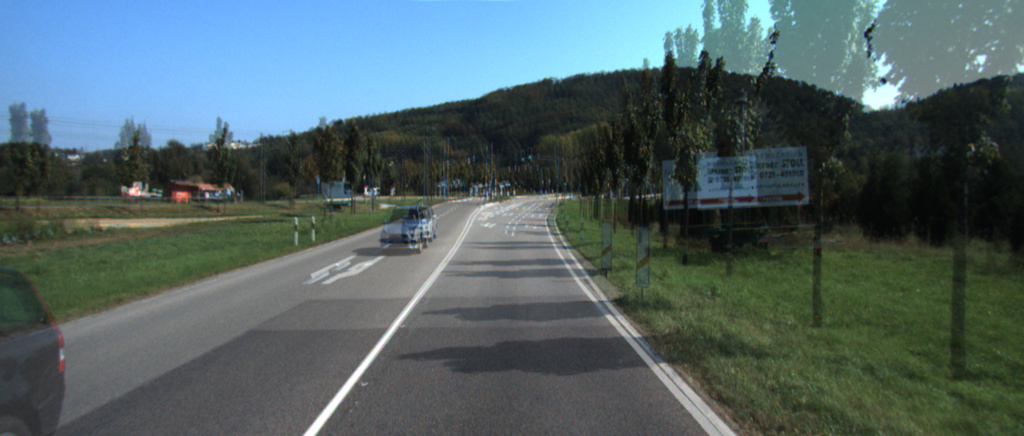}
	}%
	\centerline{
	\includegraphics[width=\interpolationresult]{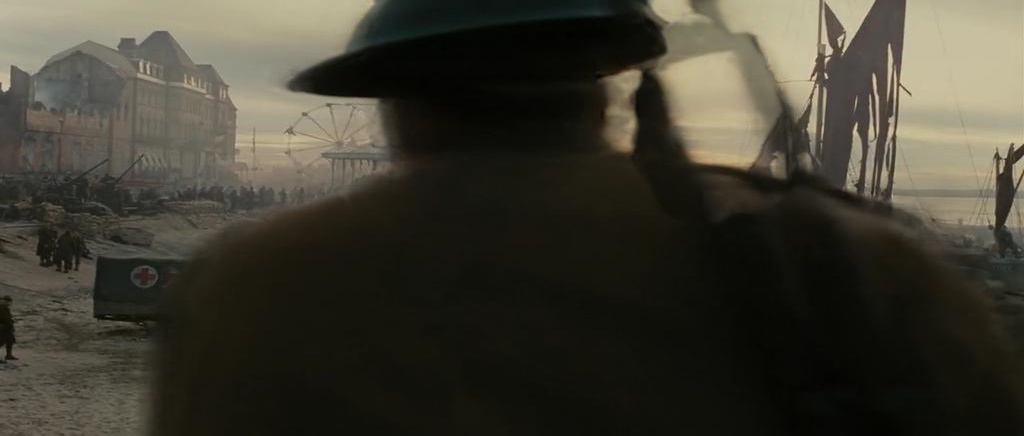}
	\includegraphics[width=\interpolationresult]{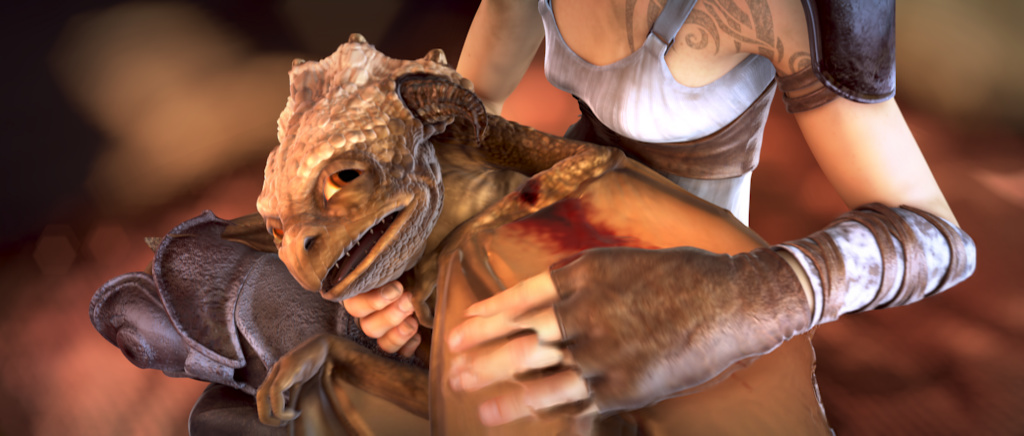}
	\includegraphics[width=\interpolationresult]{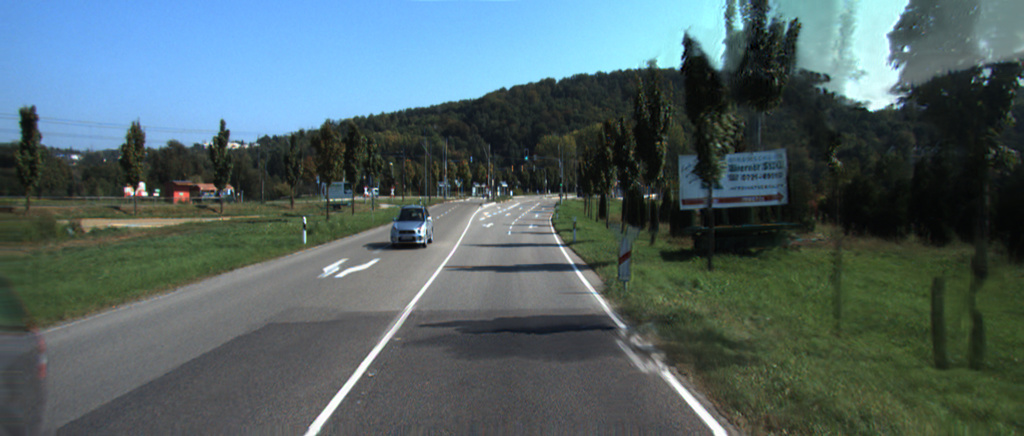}
}%
	\centerline{
	\includegraphics[width=\interpolationresult]{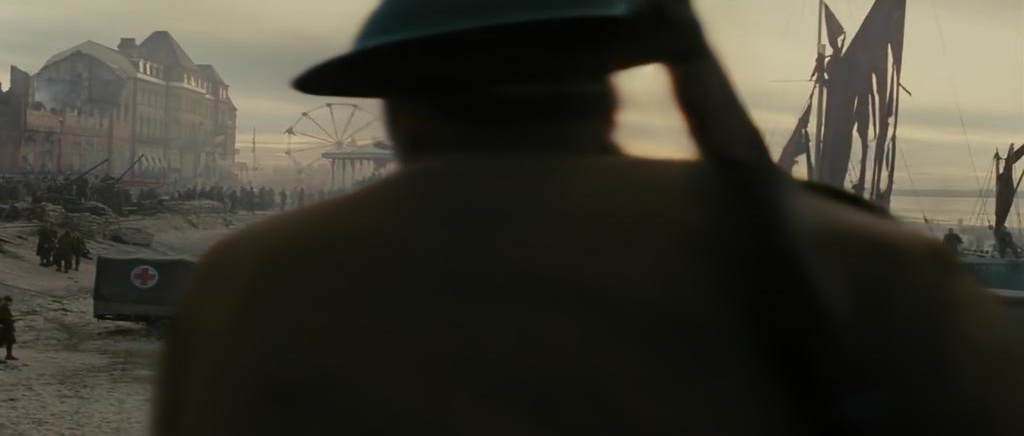}
	\includegraphics[width=\interpolationresult]{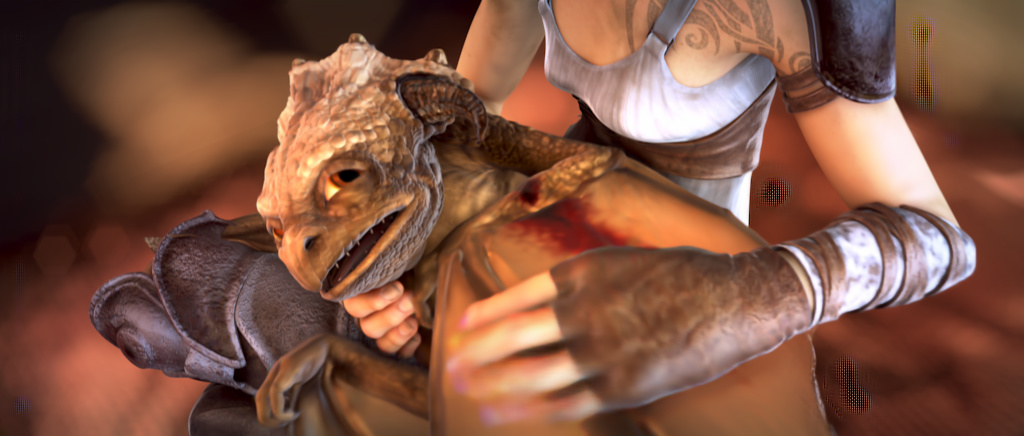}
	\includegraphics[width=\interpolationresult]{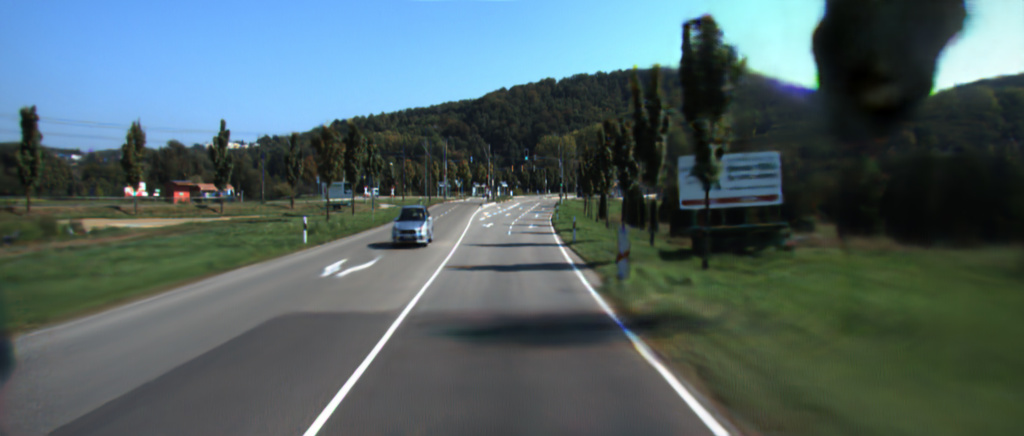}
}%
	\centerline{
	\includegraphics[width=\interpolationresult]{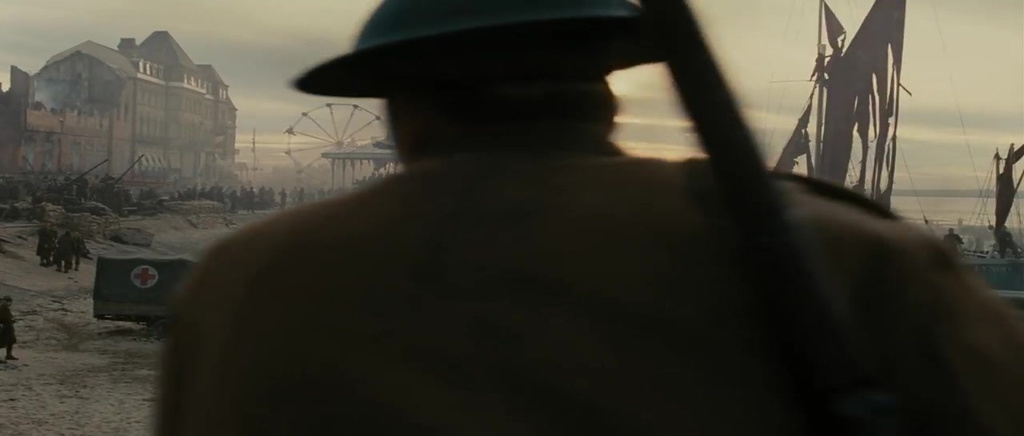}
	\includegraphics[width=\interpolationresult]{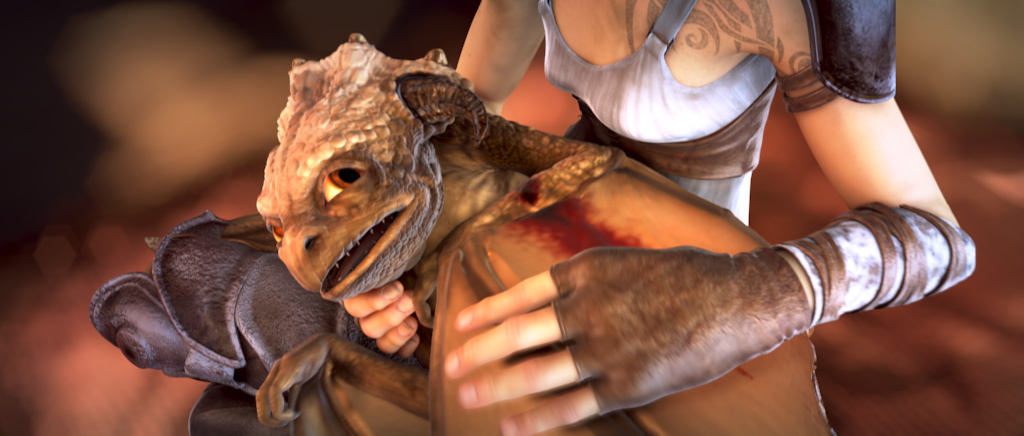}
	\includegraphics[width=\interpolationresult]{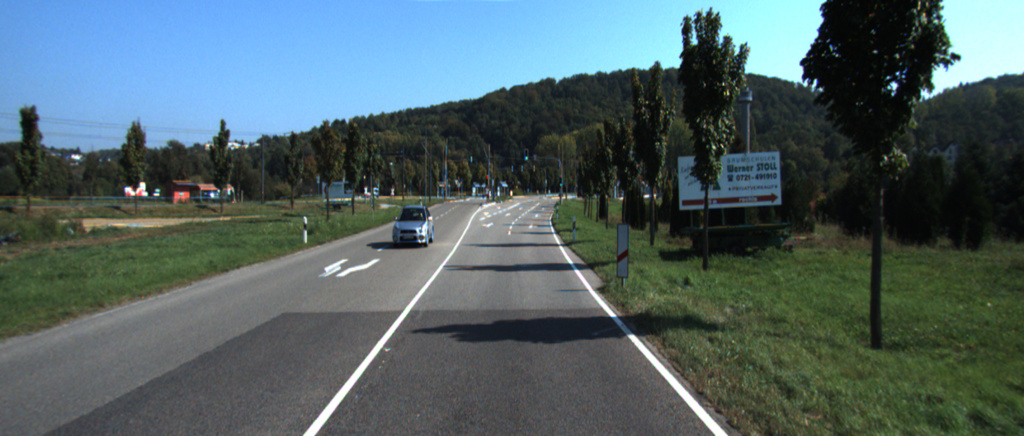}
}
	\caption{\small Visual interpolation results (unseen data). From top to bottom: linear blending; interpolation using~\cite{Niklaus:2017:AdaptiveConvolution}, $\mathcal{L}_F$ variant; interpolation using our method; ground truth. While the results from~\cite{Niklaus:2017:AdaptiveConvolution} are sharper, they produce significant artifacts, for example the tree in the right example. Our method tends to be blurrier, but captures the gist of the scene better; this is  reflected in the superior quantitative results.}
	\label{fig:interpolationresults}
\end{figure}
To evaluate the interpolation performance of our network, we compare with~\cite{Niklaus:2017:AdaptiveConvolution}, a state-of-the-art method for frame interpolation.
Unlike ours,~\cite{Niklaus:2017:AdaptiveConvolution} is specifically designed for this task; in contrast, we use a standard hourglass network.
We compute temporal interpolations for 2800 frames from natural movies, a synthetic movie (Sintel), and driving scenarios.
All frames were not previously seen in training.
To compute interpolated color frames, we simply run our network once for each input color channel.
Table~\ref{tab:interpolation} shows results on both PSNR and SSIM. For~\cite{Niklaus:2017:AdaptiveConvolution}, we report both the $\mathcal{L}_1$ and $\mathcal{L}_F$ results; according to~\cite{Niklaus:2017:AdaptiveConvolution}, the former is better suited for numerical evaluation, while the later produces better visual results.
We outperform both variants in both metrics.

\begin{table}
	\caption{\small Interpolation performance in PSNR (SSIM).}	
	\setlength{\tabcolsep}{6pt}
	\centering
	\begin{tabular}{lcccc}  
		\toprule
		& Real movie & Synthetic movie & Driving & All \\
		\midrule
		\cite{Niklaus:2017:AdaptiveConvolution}, $\mathcal{L}_1$ & 33.15 (0.915) & 25.73 (0.841) & 18.26 (0.664) & 28.80 (0.854) \\
		\cite{Niklaus:2017:AdaptiveConvolution}, $\mathcal{L}_F$ & 32.98 (0.911) & 25.44 (0.825) & 18.04 (0.631) & 28.59 (0.843) \\
		Ours & \textbf{34.68} (\textbf{0.928}) & \textbf{26.46} (\textbf{0.859}) & \textbf{19.76} (\textbf{0.710}) & \textbf{30.13} (\textbf{0.8741}) \\
		\bottomrule
	\end{tabular}
	\label{tab:interpolation}
\end{table}

Figure~\ref{fig:interpolationresults} shows quantitative results of the interpolation for all three scenarios.
Visually, the interpolation results are good.
In particular, our method can handle large displacements, as can be seen in the helmet strap in Fig~\ref{fig:interpolationresults}, first column, and the tree in Fig.~\ref{fig:interpolationresults}, third column.
The results of~\cite{Niklaus:2017:AdaptiveConvolution} tend to be sharper; however, this comes with strong artifacts visible in the second row of Fig.~\ref{fig:interpolationresults}.
Both the helmet strap and the tree are not reconstructed significantly better than when using simple linear blending (first row); our method, while slightly blurry, localizes the objects much better.

\subsection{Optical Flow}
\textbf{Results on benchmarks.}
To demonstrate the effectiveness of our method, dubbed \textit{IPFlow} for Interpolation Pretrained Flow, we test the optical flow performance on the two main benchmarks, KITTI~\cite{Geiger2013:KITTI,Menze2015GCPR} and MPI-Sintel~\cite{Butler:ECCV:Sintel}.
Since our method uses four input frames, we double the first and last frames to compute the first and last flow field within a sequence, respectively, thereby obtaining flow corresponding to all input frames.
Furthermore, like  FlowNet~\cite{Dosovitskiy:2015:FlowNet}, we perform a variational post-processing step to remove noise from our flow field.
Computing the flow on a NVIDIA M6000 GPU takes 60 ms for the CNN; the variational refinement takes 1.2 seconds.
\begin{table}
	\setlength{\tabcolsep}{6pt}
	\centering
	\caption{\small Quantitative evaluation of our method.}	
	\begin{tabular}{lcccc}  
		\toprule
		& \multicolumn{2}{c}{Sintel} & Kitti-2012 & Kitti-2015 \\
		\cmidrule(lr){2-3}
		& Clean & Final & & \\
		\midrule
		\textbf{Supervised methods} & & \\
		FlowNet2-ft~\cite{Ilg:2017:Flownet2} & \textbf{4.16} & \textbf{5.74} & \textbf{1.8} & \textbf{11.48 \%} \\
		FlowNetS+ft+v~\cite{Dosovitskiy:2015:FlowNet} & 6.16 & 7.22 & 9.1 & \\
		SpyNet+ft~\cite{Ranjan:2016:Spynet} & 6.64 & 8.36 & 4.1 & 35.07 \% \\
		\midrule
		\textbf{Un- and semisupervised methods} & & \\
		DSTFlow~\cite{Ren:2017:UnsupervisedOpticalFlowEstimation} & 10.41 & 11.28 & 12.4 & 39 \% \\
		USCNN~\cite{Ahmadi:2016:UnsupervisedCNNForMotion} & & 8.88 &  & \\
		Semi-GAN~\cite{Lai:2017:SemiSupervisedFlow} & 6.27 & 7.31 & 6.8 & 31.01 \% \\
		UnFlow-CSS~\cite{Meister:2017:UnFlow} & 9.38 & 10.22 & \textbf{1.7} & \textbf{11.11 \%} \\
		IPFlow (ours) & \textbf{5.95} & \textbf{6.59} & 3.5 & 29.54 \% \\
		IPFlow-Scratch & 8.35 & 8.87 & \\
		\bottomrule
	\end{tabular}
	\label{tab:numbers}
\end{table}

\newcommand{\resultwidth}{0.22\textwidth}
\newcommand{\resultkittiwidth}{0.31\textwidth}
\begin{figure*}
	\centering
	\centerline{%
		\includegraphics[width=\resultwidth]{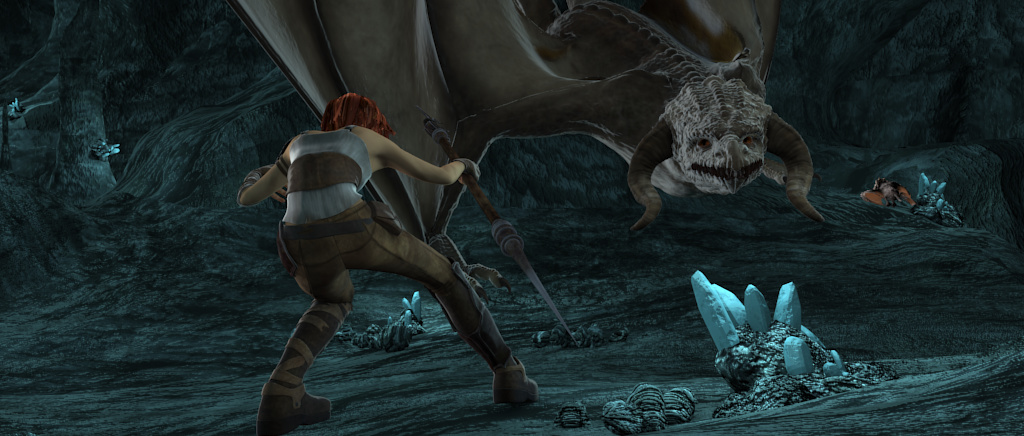}%
		\includegraphics[width=\resultwidth]{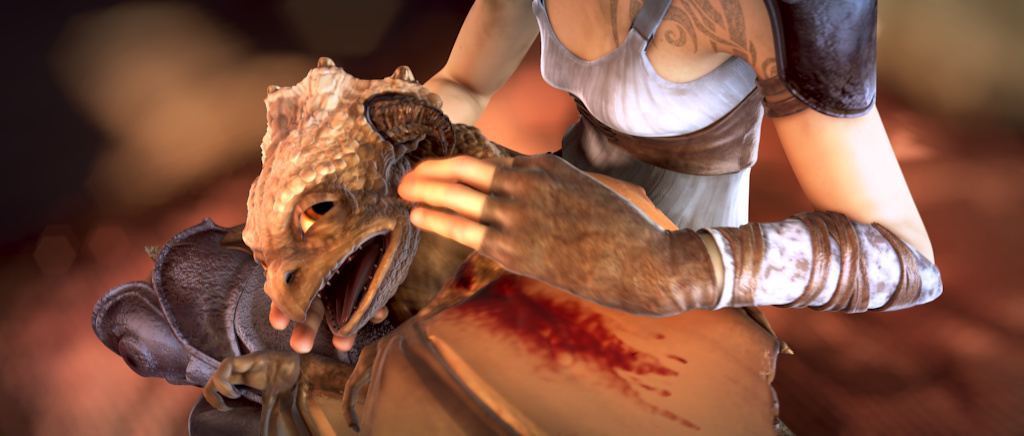}%
		\includegraphics[width=\resultwidth]{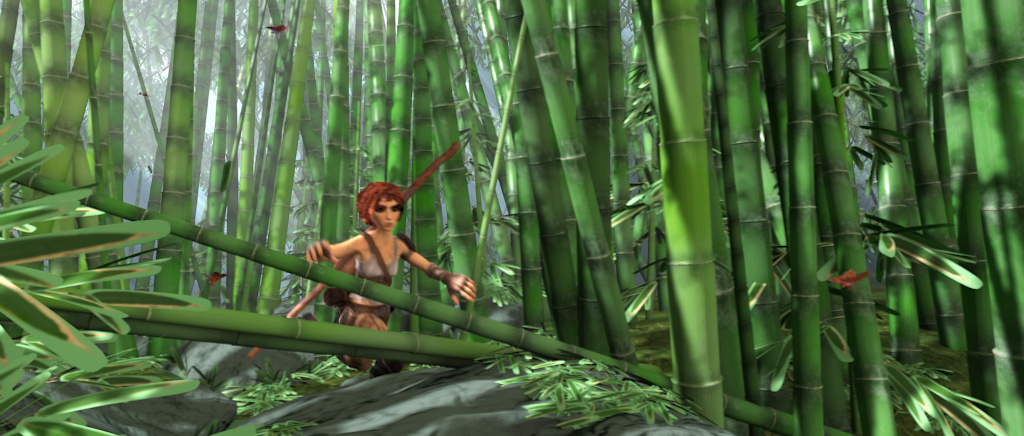}%
		\includegraphics[width=\resultkittiwidth]{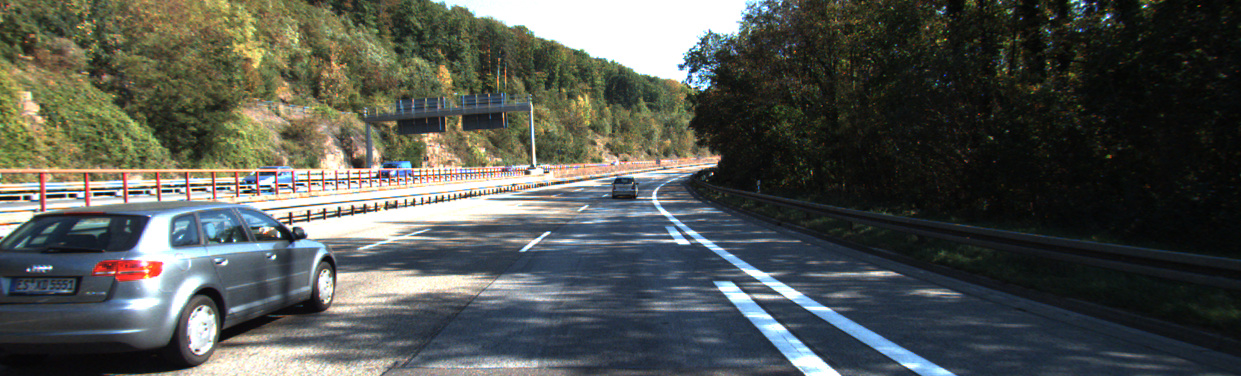}%
	}
	\centerline{%
		\includegraphics[width=\resultwidth]{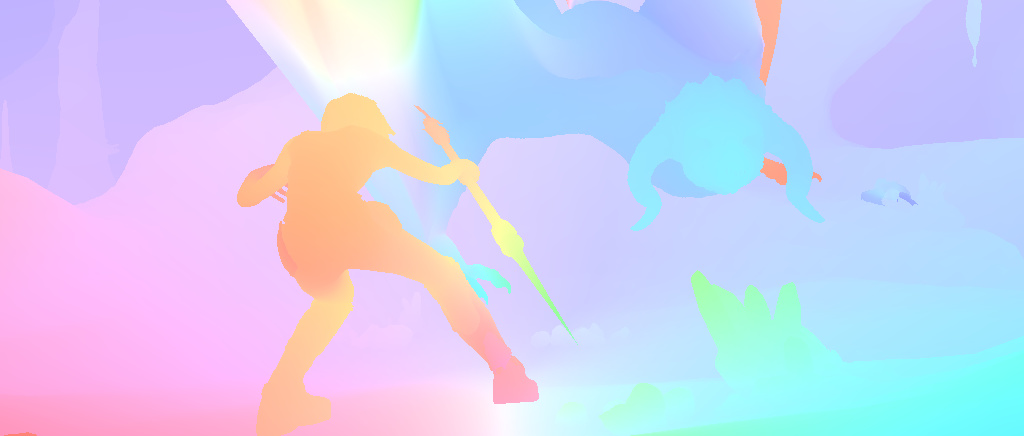}%
		\includegraphics[width=\resultwidth]{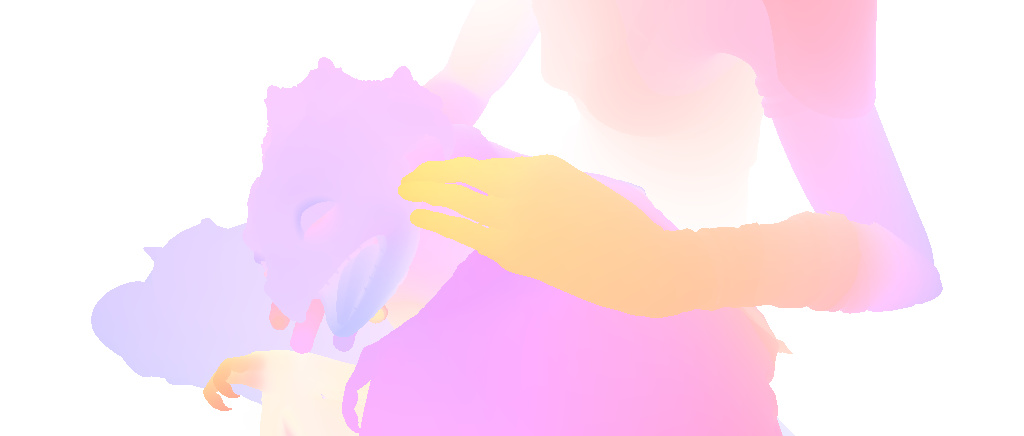}%
		\includegraphics[width=\resultwidth]{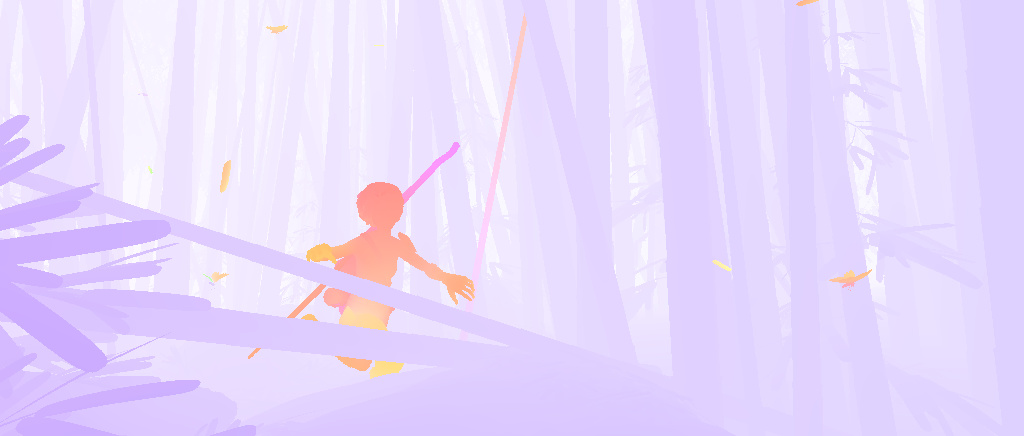}%
		\includegraphics[width=\resultkittiwidth]{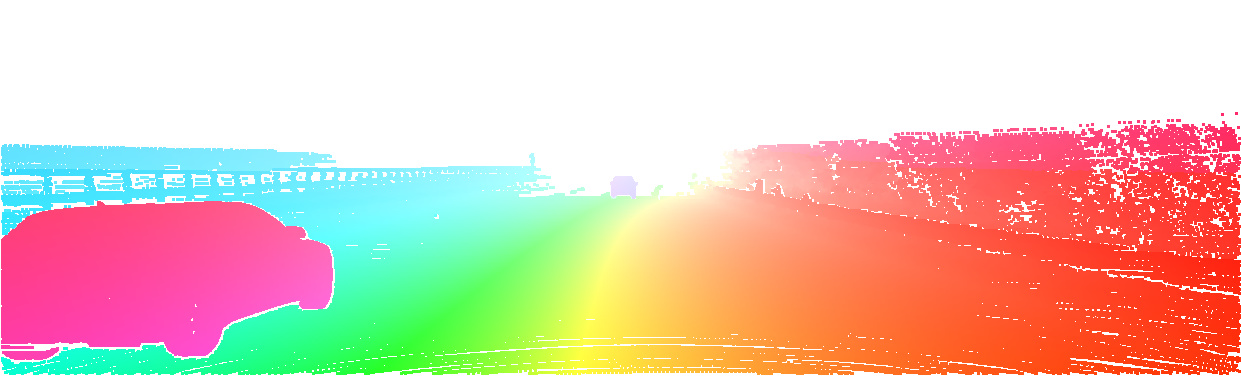}%
	}
	\centerline{%
		\includegraphics[width=\resultwidth]{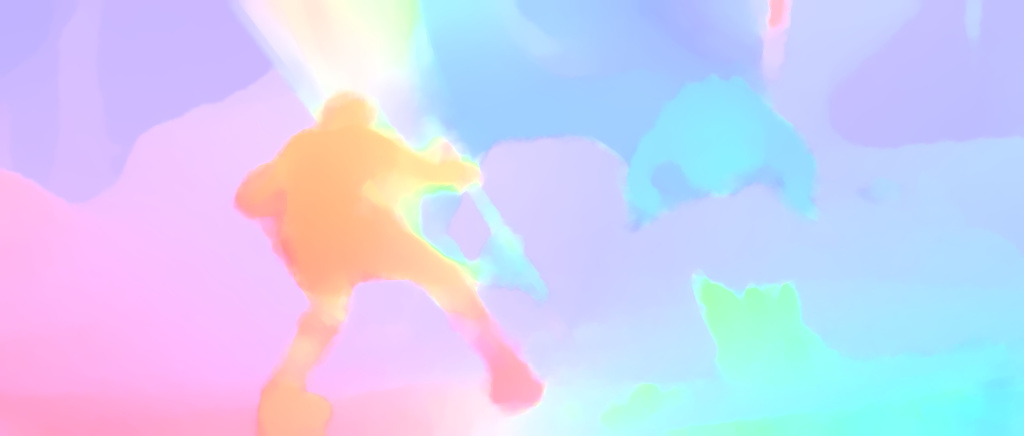}%
		\includegraphics[width=\resultwidth]{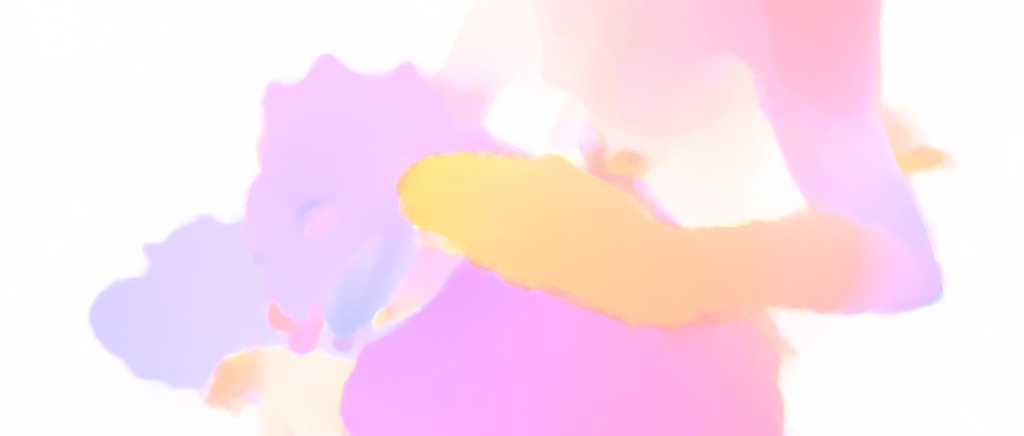}%
		\includegraphics[width=\resultwidth]{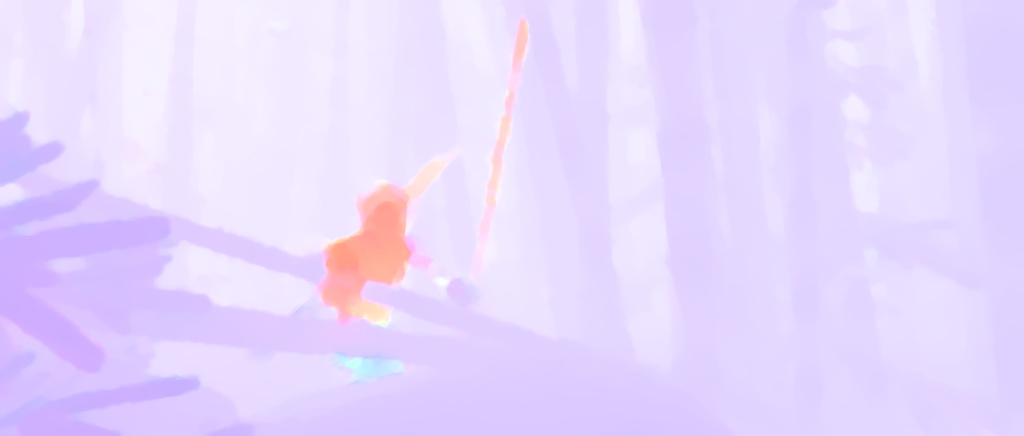}%
		\includegraphics[width=\resultkittiwidth]{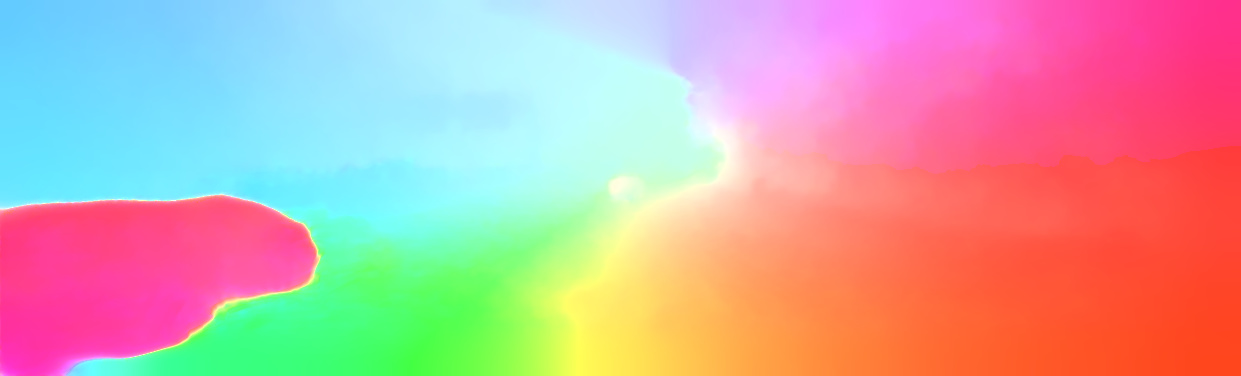}%
	}
	\centerline{%
		\includegraphics[width=\resultwidth]{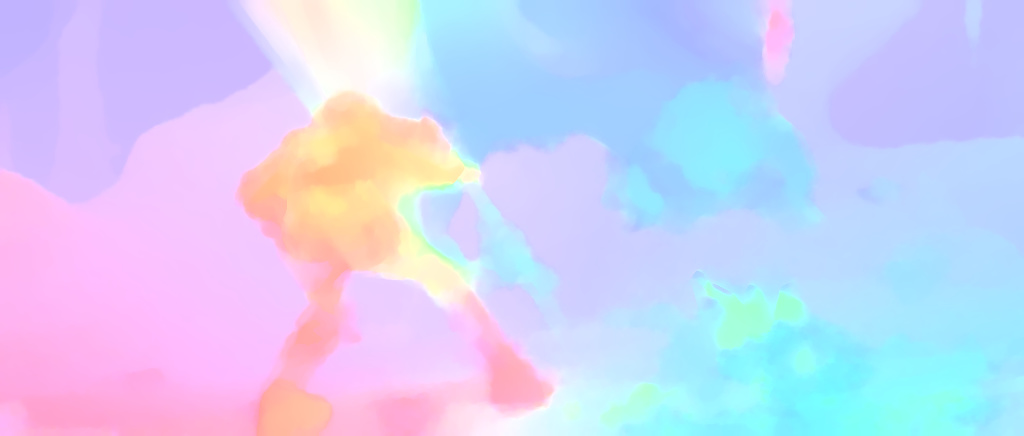}%
		\includegraphics[width=\resultwidth]{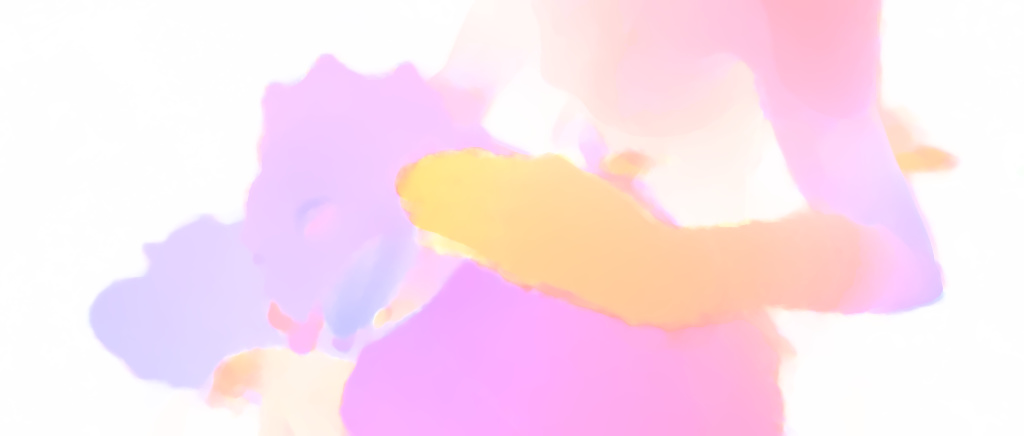}%
		\includegraphics[width=\resultwidth]{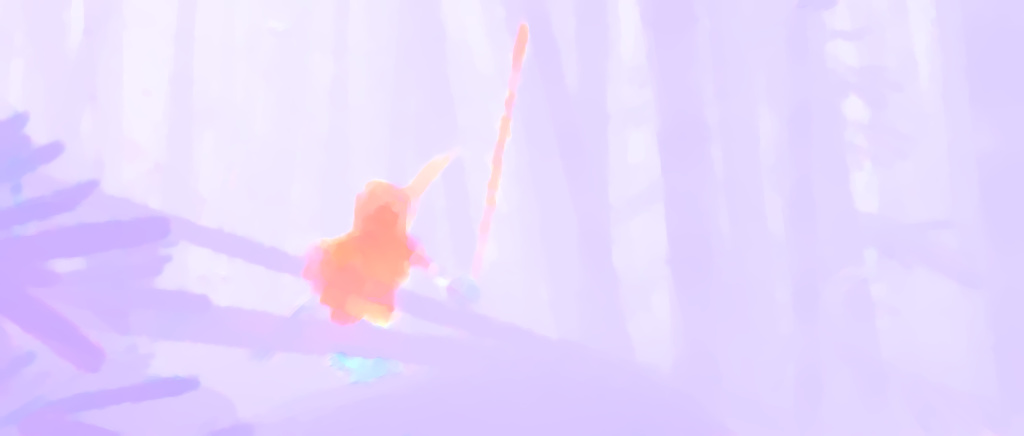}%
		\includegraphics[width=\resultkittiwidth]{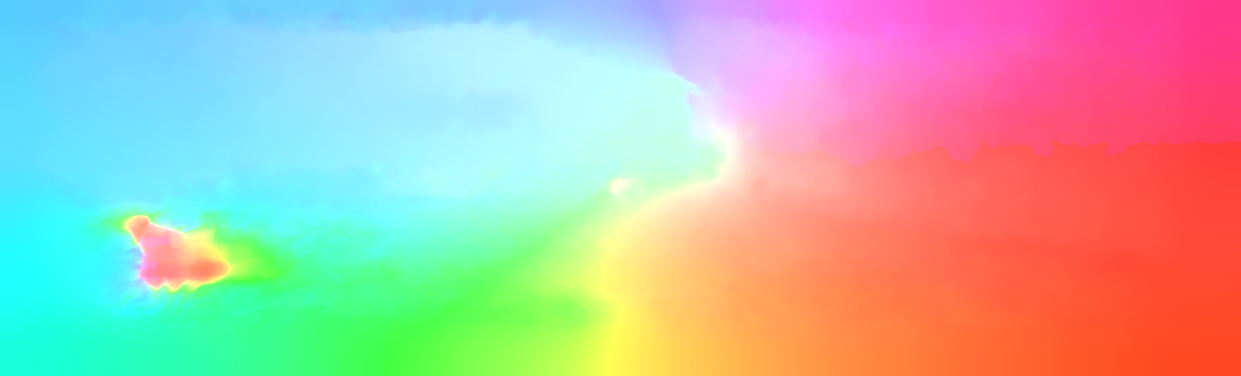}%
	}
	\caption{\small Visual results. From top to bottom: Input image; Ground truth flow; Result of IPFlow; Training from scratch. The flow computed using pure training from scratch is reasonable, but using pre-training yields significantly better optical flow maps.}
	\label{fig:results}
\end{figure*}
Table~\ref{tab:numbers} shows the errors on the unseen test sets (average endpoint error for Sintel and KITTI-2012, \texttt{Fl-All} for KITTI-2015); Figure~\ref{fig:results} shows qualitative results.
While we do not quite reach the same performance as more complicated architectures such as FlowNet2~\cite{Ilg:2017:Flownet2} or UnFlow-CSS~\cite{Meister:2017:UnFlow}\footnote{For UnFlow, test set results are only available for the -CSS variant, which is based on a FlowNet2 architecture. The simpler UnFlow-C is not evaluated on the test sets.}, on all datasets we outperform methods which are based on simple architectures comparable to ours, including those that were trained with large amounts of annotated ground truth data (FlowNetS+ft+v~\cite{Dosovitskiy:2015:FlowNet} and SpyNet~\cite{Ranjan:2016:Spynet}).
For these simple architectures, pre-training using a slow-motion task is hence superior to pre-training using synthetic, annotated optical flow data.
UnFlow-CSS is the only method outperforming ours on KITTI that does not require large amounts of annotated frames; yet, they use a considerably more complicated architecture and only achieve state-of-the-art results in driving scenarios and not on Sintel.

\textbf{Performance without pretraining.}
To evaluate whether the Sintel training data might be enough to learn optical flow by itself, we also tried training our network from scratch.
We test two training schedules, first using the same learning parameters as for the fine-tuning, and second the well-established \textit{s\_short} schedule from~\cite{Ilg:2017:Flownet2}.
As shown in Fig.~\ref{fig:scratch}, the network is able to learn to compute optical flow even without pre-training, and benefits from using the \textit{s\_short} schedule\footnote{For fine-tuning after pretraining, \textit{s\_short} gives higher errors than our schedule.}.
However, at convergence the error of the network without pre-training on unseen validation data is around 50 \% higher.
This is also visible in Table~\ref{tab:numbers}, where \textit{IPFlow-Scratch} denotes the training from scratch using~\textit{s\_short}; again, the errors are considerably higher.
Thus, important properties of motion must have been learned during the unsupervised pretraining phase.
\newcommand{\graphwidthcomparison}{0.6\textwidth}
\begin{figure}[h]
	\centering
	\includegraphics[width=\graphwidthcomparison]{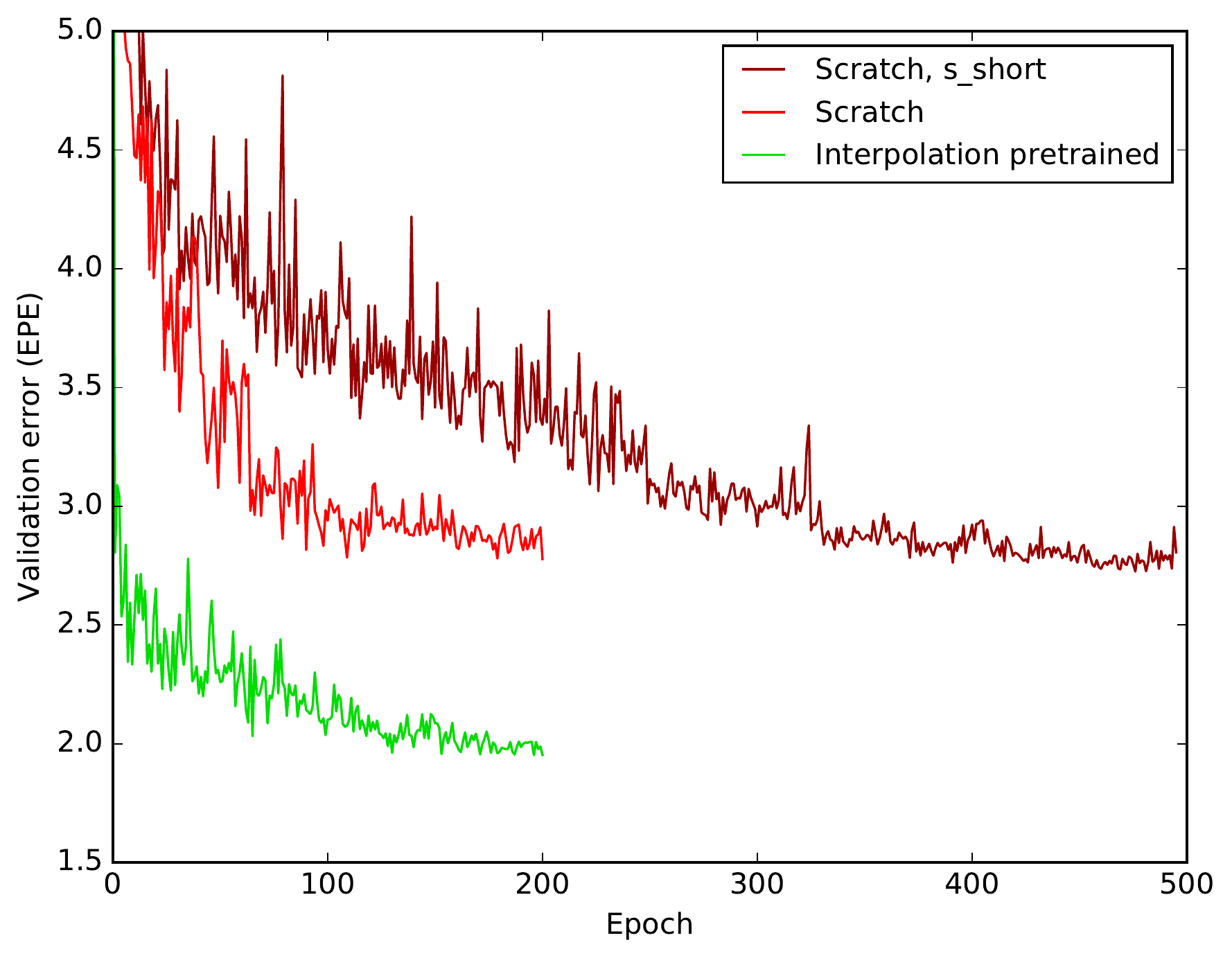}
	\caption{\small Using interpolation as pre-training, the network learns to adapt to optical flow. Flow can also be learned without pre-training, but in this case the error is 50 \% higher.}
	\label{fig:scratch}
\end{figure}

\textbf{Using a low number of fine-tuning frames.}
As we just showed, using only the training set and no fine-tuning results in significantly worse performance; pre-training from interpolation is clearly beneficial.
However, this now points to another, related question: Once we have a pre-trained network, how much annotated training data is actually required to achieve good performance?
In other words, does pre-training free us from having to annotate or generate thousands of ground truth optical flow frames, and if so, how large is this effect?

\newcommand{\graphwidthsubsampling}{0.6\textwidth}
\begin{figure}[h]
	\centering
	\includegraphics[width=\graphwidthsubsampling]{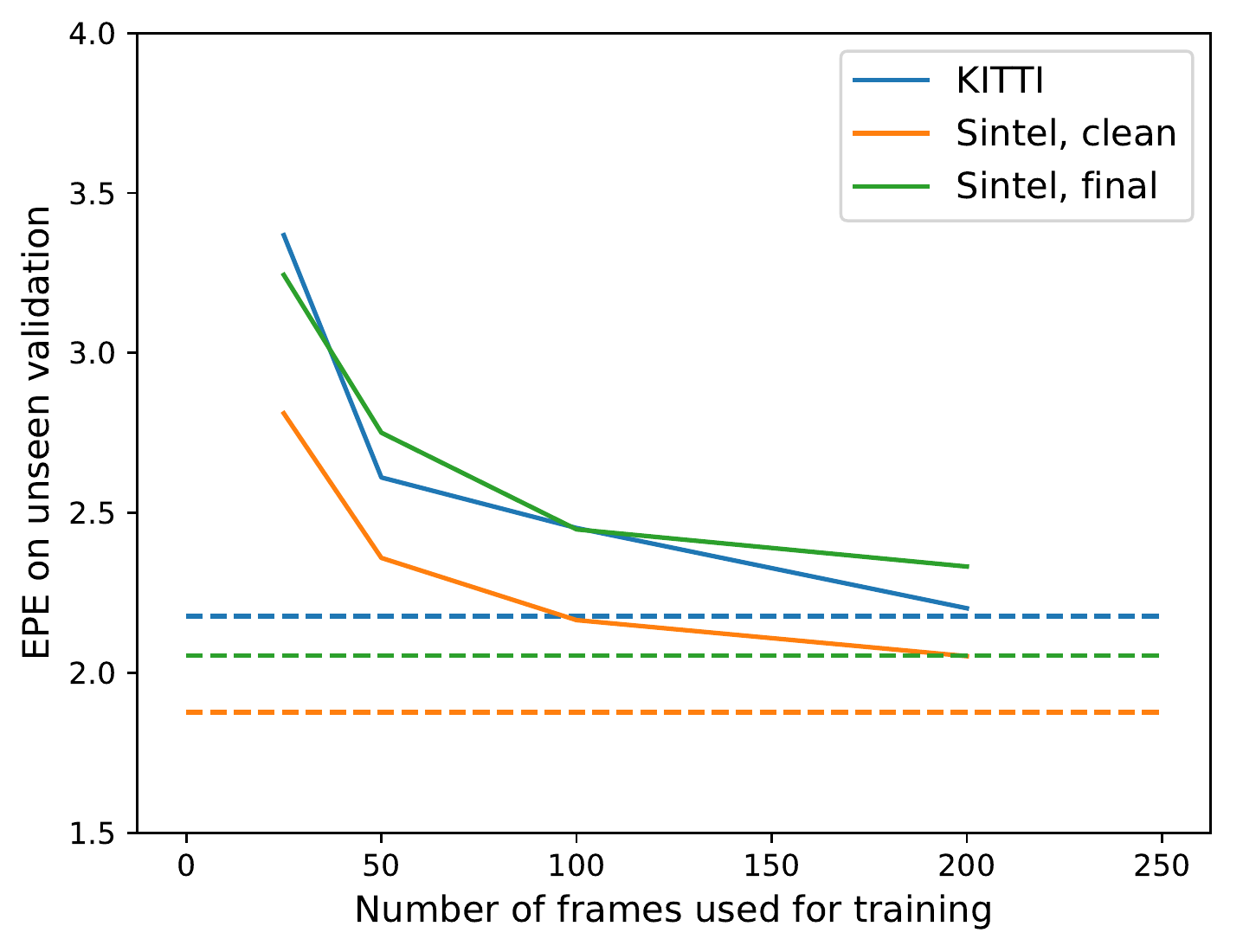}
	\caption{\small Fine-tuning with a small amount of frames. With only 100 frames, the performance on all validation sets gets within $0.5px$ EPE of the optimal performance. The dashed lines show the performance when using the full training set for each dataset.}
	\label{fig:epe_subsampling}
\end{figure}

We tested this question by repeating the finetuning procedure using only a small amount (25-200) of randomly chosen frames from the respective training sets.
Since the scenario for using very few annotated frames points to application-specific optical flow (for example, flow specifically for driving), we perform this experiment separately for different datasets, KITTI (containing both KITTI-2015~\cite{Menze2015GCPR} and KITTI-2012~\cite{Geiger2013:KITTI}), Sintel (clean) and Sintel (final).
All trained networks are tested on the same validation set for the respective dataset, and we repeated the experiment three times and averaged the results.

Figure~\ref{fig:epe_subsampling} shows the results.
While using only 25 frames is generally not enough to estimate good optical flow, the performance quickly improves with the number of available training frames.
After seeing only 100 training frames, for all datasets the performance is within 0.5 px EPE of the optimal performance achievable when using the full training sets for the respective dataset.
This shows that a interpolation-pretrained network such as the one presented here can be quickly and easily tuned for computing flow for a specific application, and does not require a large amount of annotated ground truth flow fields.
	\section{Conclusion}
In this work, we have demonstrated that a network trained for the task of interpolation does learn about motion in the world.
However, this is only true for image regions containing sufficient texture for the photometric error to be meaningful.
We have shown that, using  a simple fine-tuning procedure, the network can be taught to (a) fill in untextured regions and (b) to output optical flow directly, making it more efficient than comparable, previous work~\cite{Long:2016:LearningImageMatching}.
In particular, we have shown that only a small number of annotated ground truth optical flow frames is sufficient to reach comparable performance to large datasets; this provides the user of our algorithm with the choice of either increasing the accuracy of the optical flow estimation, or to require only a low number of annotated ground truth frames.
Demonstrating the importance of pre-training, we have shown that the same network without the interpolation pre-training performs significantly worse; our network also outperforms all other methods with comparable architectures, regardless whether they were trained using full supervision or not.
As a side effect, we have demonstrated that, given enough and sufficiently varied training data, even a simple generic network architecture outperforms a specialized architecture for frame interpolation.

Our work suggests several directions for future work.
First, it shows the usefulness of this simple proxy task for correspondence estimation.
In the analysis of still images, however, we often use a proxy task that requires some \textit{semantic} understanding of the scene, in the hope of arriving at internal representations of the image that mirror the semantic content.
As video analysis moves away from the pixels and more towards higher-level understanding, finding such proxy tasks for video remains an open problem.
Second, even when staying with the problem of optical flow estimation, we saw that optimized pipelines together with synthetic training data still outperform our method.
We believe, however, that even those algorithms could benefit from using a pre-training such as the one described here; utilizing it to achieve true state-of-the-art performance on optical flow remains for future work.

Lastly, the promise of unsupervised methods is that they scale with the amount of data, and that showing more video to a method like ours would lead to better results.
Testing how good an interpolation (and the subsequent flow estimation) method can get by simply watching more and more video remains to be seen.

\textbf{Acknowledgements \& Disclosure.} JW was supported by the Max Planck ETH Center for Learning Systems. MJB has received research funding from Intel, Nvidia, Adobe, Facebook, and Amazon. While MJB is a part-time employee of Amazon, this
research was performed solely at, and funded solely by, MPI.

	{\small
		\bibliographystyle{splncs04}
		\bibliography{interpolation2}
	}

\end{document}